\documentclass{article}

\usepackage{arxiv}

\usepackage[utf8]{inputenc} 
\usepackage[T1]{fontenc}    
\usepackage{hyperref}       
\usepackage{url}            
\usepackage{booktabs}       
\usepackage{amsfonts}       
\usepackage{nicefrac}       
\usepackage{microtype}      
\usepackage{lipsum}		
\usepackage{graphicx}
\usepackage[square,sort,comma,numbers]{natbib}
\usepackage{doi}
\usepackage{multirow} 
\usepackage{makecell}

\title{Fusion of Diffusion Weighted MRI and Clinical Data for Predicting Functional Outcome after Acute Ischemic Stroke with Deep Contrastive Learning}

\author{\hspace{1mm}Chia-Ling~Tsai\thanks{Equal contribution.\\ \textsuperscript{1}Queens College of the City University of NY, USA.\\
\textsuperscript{2}National Chung Cheng University, Taiwan\\
\textsuperscript{3}Ditmanson Medical Foundation Chia-Yi Christian Hospital, Taiwan\\
Correspondence to: \texttt{ctsai@qc.cuny.edu.}}\hspace{1.5mm}\textsuperscript{1} \\
	\And
	\hspace{1mm}Hui-Yun~Su\textsuperscript{*2} \\
	\And
    \hspace{1mm}Shen-Feng~Sung\textsuperscript{3}\\
    \And
    \hspace{1mm}Wei-Yang~Lin\textsuperscript{2}\\
    \And
    \hspace{1mm}Ying-Ying~Su\textsuperscript{3}\\
    \And
    \hspace{1mm}Tzu-Hsien~Yang\textsuperscript{3}\\
    \And
    \hspace{1mm}Man-Lin~Mai\textsuperscript{2}
}

\renewcommand{\headeright}{Technical Report}

\hypersetup{
pdftitle={Fusion of Diffusion Weighted MRI and Clinical Data for Predicting Functional Outcome after Acute Ischemic Stroke with Deep Contrastive Learning},
pdfsubject={q-bio.NC, q-bio.QM},
pdfauthor={Chia-Ling~Tsai,Hui-Yun~Su, Shen-Feng~Sung, Wei-Yang~Lin, Ying-Ying~Su,
Tzu-Hsien~Yang, Man-Lin~Mai},
pdfkeywords={Acute ischemic stroke, Diffusion-weighted MRI, Hierarchical multimodal fusion, Multimodal representation learning, Stroke prognostic model}
}
\begin{document}
\maketitle

\begin{abstract}
Stroke is a common disabling neurological condition that affects about one-quarter of the adult population over age 25; more than half of patients still have poor outcomes, such as permanent functional dependence or even death, after the onset of acute stroke. The aim of this study is to investigate the efficacy of diffusion-weighted MRI modalities combining with structured health profile on predicting the functional outcome to facilitate early intervention. A deep fusion learning network is proposed with two-stage training: the first stage focuses on cross-modality representation learning and the second stage on classification. Supervised contrastive learning is exploited to learn discriminative features that separate the two classes of patients from embeddings of individual modalities and from the fused multimodal embedding. The network takes as the input DWI and ADC images, and structured health profile data. The outcome is the prediction of the patient needing long-term care at 3 months after the onset of stroke. Trained and evaluated with a dataset of 3297 patients, our proposed fusion model achieves 0.87, 0.80 and 80.45\% for AUC, F1-score and accuracy, respectively, outperforming existing models that consolidate both imaging and structured data in the medical domain. If trained with comprehensive clinical variables, including NIHSS and comorbidities, the gain from images on making accurate prediction is not considered substantial, but significant. However, diffusion-weighted MRI can replace NIHSS to achieve comparable level of accuracy combining with other readily available clinical variables for better generalization.  
\end{abstract}


\section{Introduction}
Stroke is a major cause of acquired long-term disability~\cite{r1,r2} and is one of the major causes of death and disability worldwide~\cite{feigin2021global}. This disabling neurological condition affects about one-quarter of the adult population over age 25, with a rising incidence in young people. Even with advanced acute treatment of strokes, more than half of patients who have had strokes still have poor outcomes, such as permanent functional dependence or even death. 

The long-term functional outcome of a stroke patient is measured by the modified Rankin Scale (mRS), which grades the degree of disability in daily activities, ranging from 0 for no symptoms to 6 for death. Studies have shown that physicians specialized in stroke treatment can only achieve an overall accuracy of 16.9\% in predicting long-term disability or death~\cite{r4,r5}. However, an accurate prognostic risk model for mRS plays a crucial role in post-acute care to facilitate shared decision-making and to mitigate the mental and financial stress of the patients and their families for the long-term care.   

\begin{figure}[t]
\centering
    \includegraphics[width=0.7\textwidth]{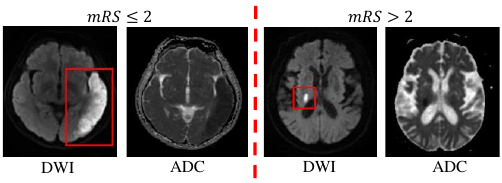}
\caption{MRI scans of brains after stroke. The image pair on the left comes from a patient with good recovery at 3 months after stroke (mRS$\leq$2), whereas the pair on the right comes from a patient having permanent disability (mRS$>$2). The red box highlights the affected region of the brain. 
} \label{fig2}

\end{figure}

Diffusion-weighted MRI imaging is a widely used modality for acute ischemic stroke (AIS) diagnosis: hyperacute lesions and very small ischemic lesions can be more easily detected, comparing to using brain CT and conventional MRI sequences~\cite{c2}. When Apparent Diffusion Coefficient (ADC) and Diffusion Weighted Image (DWI) are used jointly, the evolution of the lesion of AIS can be more accurately identified\cite{c1}. Yet, it can still be challenging for physicians to make accurate prognoses given the diffusion images at the time of stroke onset. Figure~\ref{fig2} shows two cases of stroke that can easily mislead the physicians to make incorrect prognoses due to the extent of the lesion.

Earlier research for predicting mRS after stroke mainly rely on structured data, i.e. patient-specific clinical variables, using traditional machine learning approaches, such as logistic regression~\cite{monteiro2018using,heo2019machine,li2020predicting}, random forest~\cite{monteiro2018using,heo2019machine,lin2020evaluation,alaka2020functional,li2020predicting}, support vector machine~\cite{monteiro2018using,lin2020evaluation,alaka2020functional,li2020predicting}, and {XGB}oost~\cite{monteiro2018using,li2020predicting}. Such methods can have high dependence on human-interpreted information, such as stroke severity assessment, which can vary based on the expertise of the physicians, leading to lower generalizability of the models in real clinical settings. 

Over the last decade, deep learning (DL) has shown promising success in numerous applications in medical image analysis, such as cancer screening and tumor segmentation~\cite{r6}. However, research applying DL in the area of ischemic stroke prognosis for predicting mRS after stroke remains limited in the literature. Most attempts can be classified into two categories based on the type of data used: brain image data (unstructured data), and heterogeneous data (combining both structured and unstructured data).
The former often uses classic deep learning models, including convolutional neural network (CNN) ~\cite{lai2022using}, siamese network~\cite{osama2020predicting} for cross-modality parameter sharing, and autoencoder for feature learning~\cite{hilbert2019data}. All studies suggest little benefit of using imaging alone for 3-month functional outcome prediction.

Risk models using heterogeneous data have the advantage over models using only structured or unstructured data, as both image-embedded information and patient medical history are considered for making the prediction~\cite{zihni2022moving}. Samak \emph{et al}.~\cite{samak2020prediction} exploit attention mechanism to capture channel and spatial information in the 3D image, allowing the model to focus more on information in specific regions of the image. Structured features are generated using multilayer perceptron (MLP) and concatenated with CNN image features. The combined features go through a fully connected (FC) layer to produce the final output. Simpler models are proposed by Zihni \emph{et al}.~\cite{zihni2020multimodal} and Bacchi \emph{et al}.~\cite{bacchi2020deep} as a combination of 3D CNN and MLP to process image data and structured data, respectively. They only differ in the number of layers inside CNN and MLP. Feature concatenation also takes place before the FC layer for classification. Instead of using 3D CNN, Hatami \emph{et al}.\cite{hatami2022cnn} process a 3D image as a sequence of 2D images using a combination 2D CNN and a recurrent neural network LSTM, and the final output is weighted by one single structured attribute, such as age, to enhance the outcome. Overall, multi-modal networks have slightly better predictive performance compared to the model trained on only clinical variables, but improvement over the image-trained CNN models is prominent. There is a need to explore alternative fusion approaches to achieve better performance. 

Models designed for heterogeneous data have also been applied to other medical problems. Hsu \emph{et al}. ~\cite{hsu2021deep} apply the same approach as~\cite{zihni2020multimodal} and~\cite{bacchi2020deep} to fundus images and structured data to detect diabetic retinopathy, but Inception v4 is used as the backbone instead of a vanilla 2D CNN. Following a similar approach, Huang \emph{et al}.~\cite{huang2020multimodal} compute the average of the predicted probabilities from the 3D CNN and MLP, respectively, as the final prediction for the detection of pulmonary embolism. They also experimented with various fusion methods with higher model complexity but produced less desirable outcome. Wood \emph{et al}.~\cite{wood2022deep} design a model that concatenates the age with CNN image features as input to a FC layer to produce the final output to detect abnormal MRI brain scans. Qiu \emph{et al}.~\cite{qiu2018fusion} apply hierarchical voting which first votes on CNN outputs for individual MRI slices and then votes on the final CNN vote and multiple MLP outputs for the structured data to detect cognitive impairment in patients. In summary, all aforementioned models perform late fusion of multiple modalities by either concatenation/averaging of features or multiplication of the outcomes from individual modalities. In all cases, cross-modality learning for discriminative features is not facilitated to take into consideration the commonality between modalities and the information granularity of each modality. 

\begin{figure}[t]
\centering    \includegraphics[width=0.8\textwidth]{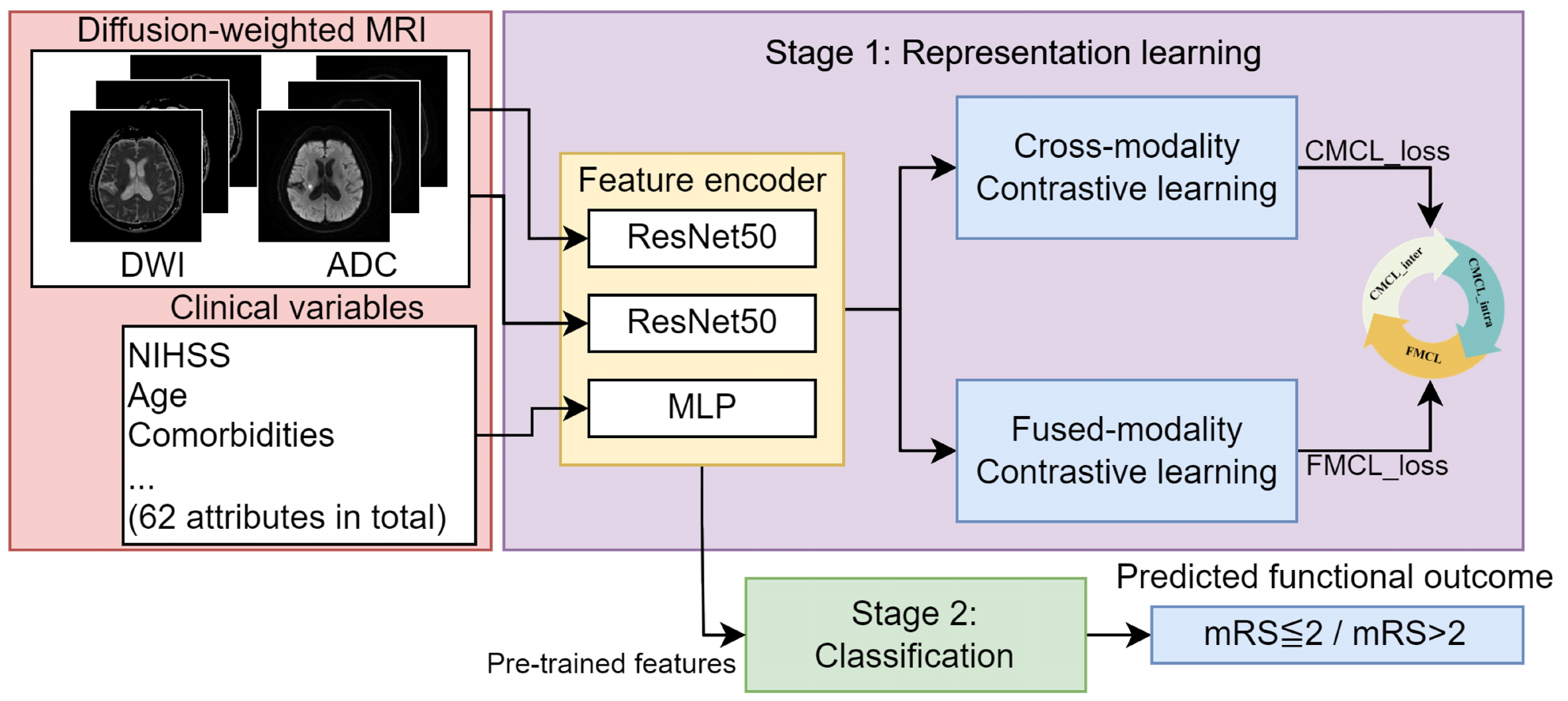}
\caption{Overview of the proposed multimodal prognostication framework for acute ischemic stroke. } 
\label{summary}

\end{figure}

In this study, we investigate the efficacy of combining diffusion-weighted MRI imaging with clinical variables for predicting the 3-month functional outcome (mRS$>2$) with a cross-modality fusion model. Our  proposed model performs representation learning to learn discriminative features cross modalities that separate the two classes of patients from embeddings of individual modalities and from the fused multimodal embedding. For the latter, data fusion is performed in a hierarchical fashion to ensure equal contributions from both fine-grained and coarse-grained representations. Figure~\ref{summary} provides a visual summary of our proposed multimodal prognostication framework.

\section{Materials and Methods}
\subsection{Dataset}
The study protocol is independently approved by the Ditmanson Medical Foundation Chia-Yi Christian Hospital Institutional Review Board (IRB2022011). Study data are maintained with confidentiality to ensure the privacy of all participants.

Our dataset consists of ADC and DWI, both taken between 1 to 7 days post-stroke, and 62 clinical variables, including age, onset-to-admission delay, comorbidities and the National Institutes of Health Stroke Scale (NIHSS). The dataset contains 3297 patients from Ditmanson Medical Foundation Chia-Yi Christian Hospital, being hospitalized for AIS between October 2007 and September 2021 and having completed the 3-month post-stroke follow-up. Only the earliest hospitalization for each patient is considered. Table~\ref{table:dataset} gives the summary of the cohort characteristics. Images with substantial artifacts, such as blurring or dentures, are also excluded from this study. The dataset is divided into training, validation, and testing with the ratio of 6:2:2. The number of slices of an MRI volume varies between 18 and 28, depending on the shooting angle and the size of the patient's head. We select the middle 18 slices of each patient as they contain the most lesion information. 

\begin{table}[ht]
\renewcommand{\arraystretch}{1.5}
\caption{Summary of cohort characteristics. IQR: interquartile range. NIHSS: National Institutes of Health Stroke Scale.}
    \centering
    \begin{tabular}{|c||c|c|}
    \bottomrule
        \textbf{Clinical   attributes}      & \textbf{mRs=[0,2] (N=1802)} & \textbf{mRs=[3,6] (N=1495)} \\ \hline
        Median   age (IQR)                  & 67.0   (17.0)      & 75.0   (15.0)      \\ \hline
        Median   initial NIHSS (IQR)        & 4.0 (4.0)          & 8.0 (12.0)         \\ \hline
        Sex   (females/males)            & 610 / 1192         & 692 / 803          \\ \hline
        Thrombolysis   treatment (yes/no)   & 184 / 1618         & 130 / 1365         \\ \hline
        Diabetes   (yes/no)                 & 691 / 1111         & 695 / 800          \\ \hline
        Smoking   (yes/no)                  & 805 / 997          & 519 / 976          \\ \hline
        Onset-to-admission   delay (yes/no) & 1316 / 486         & 1102 / 393         \\ \hline
        Hyperlipidemia   (yes/no)           & 1082 / 720         & 804 / 691          \\ \hline
        Hypertension   (yes/no)             & 1373 / 429         & 1222 / 273         \\ \hline
        Cardiac   history (yes/no)          & 335 / 1467         & 484 / 1011         \\
    \toprule
    \end{tabular}
\label{table:dataset}
\end{table}

In term of data augmentation, random flipping, Gaussian blurring with the std in [0.1, 2.0], and random noise addition with a probability of 0.2 are applied in the first stage of training. In addition, partial masking is applied with a patch size of $32\times 32$ and masking probability of 0.5 for each patch, as proposed by He \emph{et al}.~\cite{he2022masked}, to ensure uniform mask distribution across the image. For structured data, we add dropout 0.5 to the model to mask the structured features. No data augmentation is applied in the second stage of training. 

\subsection{Data Preprocessing}
\begin{figure}[t]
\centering
\includegraphics[width=0.6\textwidth]{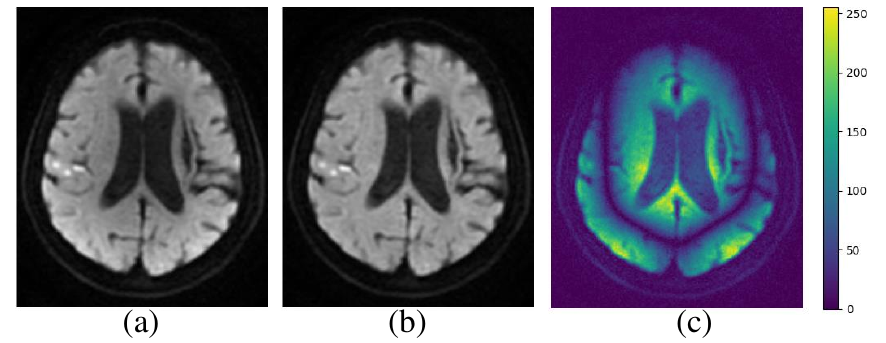}
\caption{Preprocessing for DWI. (a) Before bias correction. (b) After bias correction. (c) Bias field being removed.} 
\label{method:dpreprocess}
\end{figure}

Due to imperfection of the image acquisition process, bias field is often perceived in medical images as a smooth variation of intensity across one image. It is especially prominent in MRI because of the variation in magnetic susceptibility among tissue types. This effect causes variation in the intensity of the same tissue in different locations within the image. Bias field can greatly degrade the accuracy of many automated medical image analysis techniques~\cite{song2017review}.  To mitigate the effect of bias field in training of the model, we preprocess the ADC and DWI sequence images with N4 Bias Field Correction~\cite{tustison2010n4itk} implemented in SimpleITK\cite{lowekamp2013design}. This method is to remove the low-frequency intensity deviation to homogenize the image. As shown in Figure~\ref{method:dpreprocess}, the DWI, in both the white and gray matters, has more homogeneous brightness across the image after bias correction. 

Missing clinical attributes is a common problem in a real clinical setting. The highest missing rate is 15.4\% for a data attribute in our dataset. To resolve this issue, we experimented with several data imputation techniques, including MissForest~\cite{stekhoven2012missforest}, and settled with mode imputation for better empirical outcomes. Data imputation is applied to the input data before feeding the backbone network. 

\subsection{Cross-modality Fusion Network}
\begin{figure}[t]
\includegraphics[width=\textwidth]{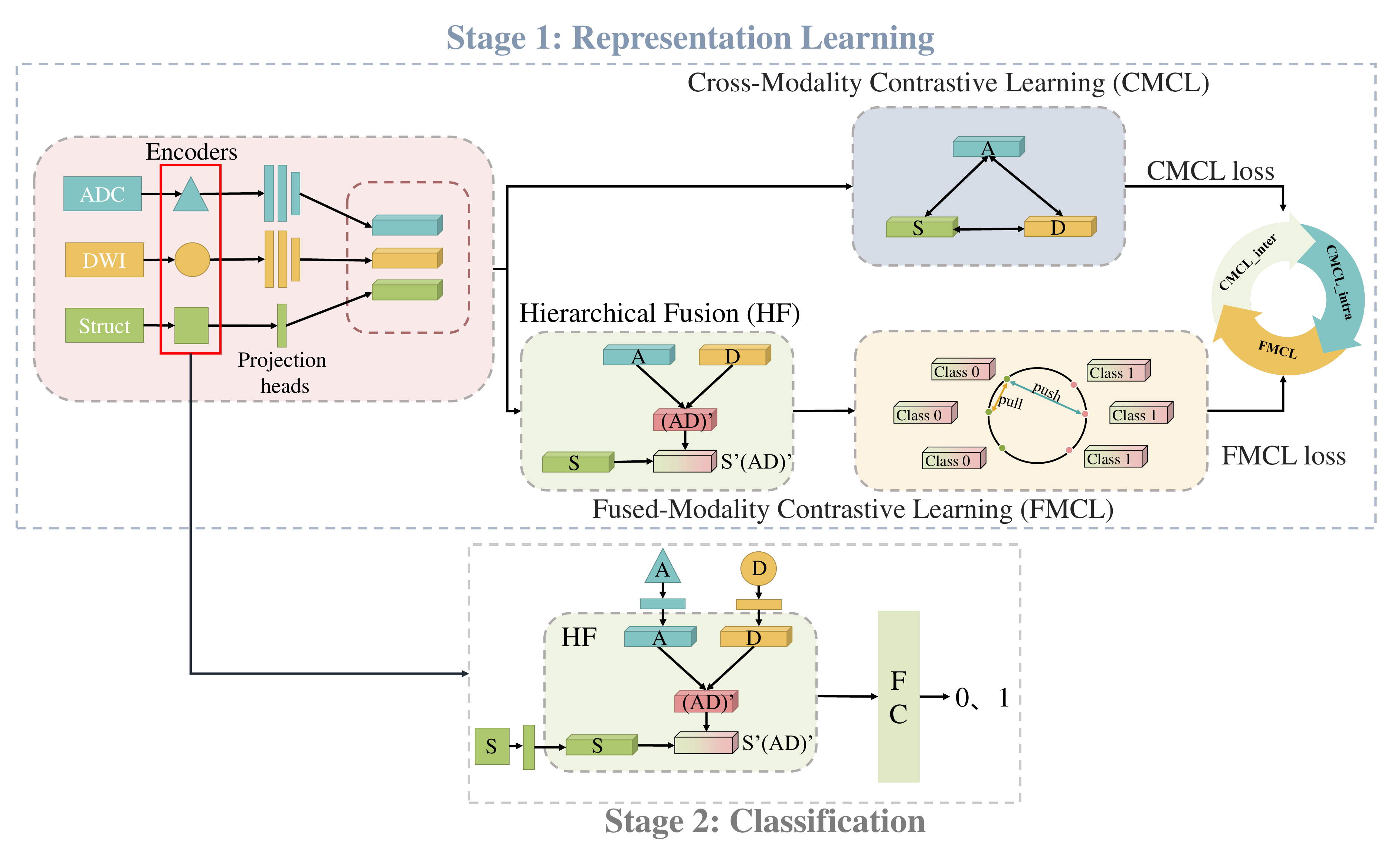}
\caption{Illustration of the proposed multimodal fusion learning network. The outcome is the prediction of the patient needing long-term care at 3 months after the onset of acute ischemic stroke. } \label{fig1}
\end{figure}

The proposed architecture consists of two training stages, as shown in Figure~\ref{fig1}. In the first stage of training, representation learning is accomplished through contrastive learning, inspired by~\cite{chen2021multimodal} for self-supervised learning. The idea of contrastive learning is to cluster together similar samples and push apart different samples in feature embedding~\cite{chen2020simple,supervisedcl}. Features are extracted using either a CNN network, which is ResNet50, for unstructured data or MLP for structured data. The same feature sets go into both cross-modality contrastive learning (CMCL) module and fused-modality contrastive learning (FMCL). The losses from the two modules are randomly selected to produce the final contrastive loss.

In the second stage of training, which is for classification, the backbone networks stay the same as in stage 1 and are initialized with parameters learned in stage 1 for more discriminative features from contrastive learning to achieve higher classification accuracy.

\subsubsection{Encoder}
In the deep learning method, an encoder is to transform an image to a latent representation to capture the important information in a condensed form. ResNet50\cite{resnet} is the backbone for encoding both DWI and ADC. The two modalities are processed separately by two networks, both initialized by ImageNet pre-trained weights. For the structured data, a MLP is used as an encoder, with 3 hidden layers---150, 100, and 60 nodes, respectively.

\subsubsection{Projection Head}
Earlier literature~\cite{chen2020simple} demonstrates the necessity to include a projection head between feature encoding and contrastive loss to learn features invariant to data augmentation. Nonlinear projection head is shown to be more effective than linear projection. In our model, we design a non-linear projection head with 3 layers with decreasing dimensions for representation learning. For the downstream task, only the layer with the lowest dimension is maintained to reduce computation without loss in performance.

\subsubsection{Cross-Modality Contrastive Learning (CMCL) Module}

Closely following SimCLR~\cite{chen2020simple}, a simplified framework for self-supervised contrastive learning with the emphasis on data augmentation, we introduce a supervised contrastive loss that incorporates labels to facilitate learning. Images of all modalities (including augmented images) from the same patient are all considered as views of an instance. 

Let $A\in\mathbb{R}^{d}$, $D\in\mathbb{R}^{d}$ and $ S \in \mathbb{R}^{d}$ 
represent the modality features at a given layer of feature extractor from ADC, DWI, and structured data, respectively. Each feature vector is in its original embedding. We calculate the loss function using a combination of modality pairs: pairs of views of the same modality, such as \{D,D'\}, where D and D' are from the same modality, and pairs of views from different modalities, such as \{A,S\}, where A and S are views of different modalities. All original and augmented views are considered. The contrastive loss for only \{D,D'\} pairs is the same as in \cite{supervisedcl}:
\begin{equation}
L_{DD}=\sum_{i\in I}\frac{-1}{\left | P(i) \right |}\sum_{p\in P(i)}\log \frac{\exp (D_{i}\cdot D'_{p}/\tau )}{\sum_{k\in K(i)}(D_{i}\cdot D'_{k}/\tau )}
\label{method:equation_cmcl_single}
\end{equation}

$P(i)\equiv \left \{ p\in K(i):\tilde{y_{p}}=\tilde{y_{i}} \right \}$, where $i$ is the anchor, $K(i)$ represents all samples except $i$ itself, $P(i)$ represents all the views of all samples of the same category as of $i$. $\tau \in \mathbb{ R}^{+}$ is the scalar temperature parameter. Similarly for $L_{AA}$, $L_{SS}$, $L_{DA}$, $L_{DS}$ and $L_{SA}$. 

$L_{intra}$ is the combined loss of all intra-modality losses from different views of the same modality for all 3 modalities:
\begin{equation}
L_{intra}=L_{AA}+L_{DD}+L_{SS}
\end{equation}

To consider views from different modalities, $ L_{inter}$ is the combined loss of inter-modality losses among the 3 modalities:
\begin{equation}
L_{inter}=L_{AD}+L_{DS}+L_{SA}
\end{equation}
 
 $L_{intra}$ can facilitate learning of discriminative features in individual modalities, while $L_{inter}$ promotes learning of strong features shared by all modalities for cross-modality learning. 

\subsubsection{Fused-Modality Contrastive Learning (FMCL) Module}

Features of different modalities are often concatenated or averaged out when combined. If multiple modalities of similar granularity in feature embedding are involved in the learning processing, such data representation may dictate the learning outcome. For this reason, we introduce multi-stage Hierarchical Fusion (HF): image feature vectors from the backbone networks are concatenated and processed by the first FC layer, the fused image features are concatenated with the features of the structured data and processed again by the second FC layer. Multi-stage fusion ensures equal weights for features of different granularities in the common feature embedding.

HF generates the common feature embedding $M$ from all three modalities. Same as for CMCL, the contrastive loss for FMCL is computed as:  

\begin{equation}
L_{FMCL}
=\sum_{i\in I}\frac{-1}{\left | P(i) \right |}\sum_{p\in P(i)}\log \frac{\exp (M_{i}\cdot M'_{p}/\tau )}{\sum_{k\in K(i)}(M_{i}\cdot M'_{k}/\tau )}
\label{method:equation_fmcl}
\end{equation}

\subsubsection{Combined loss}
In the stage of representation learning, we combine the losses computed by CMCL and FMCL, aiming to achieve cross-modality learning and multi-modality learning. By doing so, we obtain the pre-trained weights for our downstream predictive task. 
Our strategy is to update the final loss by randomly selecting only one loss from $L_{intra}$, $L_{inter}$, and $L_{FMCL}$ for each mini-batch without introducing additional hyper-parameters. 

\subsubsection{Classification}
As our ultimate goal is to predict the range of mRS at 3 months after stroke, we transfer the model parameters of the backbone networks obtained in the first stage of training for representation learning to fine-tune the model for classification. In stage 2, HF is also applied to fuse the features in the common embedding. A FC layer is added at the end to serve as the classifier, which outputs 0 for mRS$\leq$2, and 1 for mRS$>$2. To achieve this, we use cross-entropy loss as the loss function for classification.

\section{Results}

\subsection{Implementation Details}
The original image is resized from 256$\times$256 to 224$\times$224  while preserving the aspect ratio. 
The backbone networks for image modalities are ResNet50~\cite{resnet} pre-trained on ImageNet. Both ADC and DWI are 3D image modalities. We treat slices as channels and perform 2D convolution on the 3D image for computation efficiency, comparing to 3D CNN.

For stage 1, the image backbone networks are followed by a projection head, which takes an input of 2048 dimensions and projects to a feature space of 60 dimensions for image modalities. The identity function is used for the structured data for projection. 
For the hierarchical fusion, the FC layers of both stages consists of a linear layer with the ReLU activation function.

The networks are trained using an Adam optimizer with a mini-batch of 20 epochs. The learning rate is initially set to $1\times 10^{-3}$ and decreases gradually after every 10 epochs. The second stage uses a smaller learning rate of $1\times 10^{-4}$. All the experiments are implemented with the PyTorch platform and trained/tested on RTX 3090 GPU with 24GB of memory.

\subsection{Comparison with State-of-the-art Fusion Methods}

We adopt three metrics for evaluation of performance: area under curve (AUC), F1-score and accuracy (Acc). As the numbers of samples from the two classes in our dataset are imbalanced, we use the macro-F1-score to compute the average F1 score for each class. Accuracy is computed as the proportion of correct predictions. 

We compare our proposed method with methods that also predict mRS 3-month outcomes after stroke and also with methods developed for other applications using both images and structured data. To ensure fair comparison, except for~\cite{hatami2022cnn} which is limited to use only one structured attribute as the weight, we use the same set of structured attributes as in our study for all other models\footnote{\cite{qiu2018fusion} is not re-implemented because voting of the three modalities should be dominated by outcomes of two image modalities, which is empirically shown to capture less information than structured data (see Table~\ref{table:single}), resulting in worse performance.}. Same preprocessing of input data is also applied, but each model is individually adjusted following the original publication.  Table~\ref{tablesota}  shows the results of comparison.

\begin{table}[t]
\renewcommand{\arraystretch}{1.25}
\caption{Comparisons with other multimodal prediction models involving both images and structured data. (* indicates models developed for non-stroke applications.)}
\centering
\begin{tabular}{lcccccc}
    \bottomrule
     & \multicolumn{3}{c}{Validation set} & \multicolumn{3}{c}{Testing set} \\ \cmidrule(lr){2-4} \cmidrule(lr){5-7}
        Model &  AUC & F1 & Acc (\%) & AUC & F1 & Acc (\%) \\ \hline  
        Samak \emph{et al}.~\cite{samak2020prediction} & 0.8568 & 0.7890 & 79.70 & 0.8527 & 0.7778 & 78.94\\
        Bacchi \emph{et al}.~\cite{bacchi2020deep} & 0.8730 & 0.7638 & 77.88 & 0.8468 & 0.7470 & 76.52 \\ 
        Hatami \emph{et al}.~\cite{hatami2022cnn} & 0.7950 & 0.7190 & 73.18 & 0.7801 & 0.6835 & 70.30 \\
        Hsu \emph{et al}.*~\cite{hsu2021deep} & 0.8609 & 0.8071 & 81.21 & 0.8548 & 0.7693 & 77.73 \\
        Huang \emph{et al}.*~\cite{huang2020multimodal} & 0.8711 & 0.8021 & 80.45 & 0.8637 & 0.7657 & 76.97 \\ \hline
        \textbf{Ours} & \textbf{0.8863} & \textbf{0.8304} & \textbf{83.48} & \textbf{0.8703} & \textbf{0.7968} & \textbf{80.45} \\
\toprule
\end{tabular}
\label{tablesota}
\end{table}

\subsection{Ablation Study} \label{AS}
Since our architecture incorporates multiple loss functions for representation learning, we investigate different combinations of loss functions $L_{intra}$, $L_{inter}$ and $L_{FMCL}$ and fusion methods to determine the individual contributions of CMCL, FMCL, and HF modules. Additionally, models excluding HF are performed by averaging across three modalities to calculate $L_{FMCL}$. The results are shown in Table~\ref{tableablation}. 

Our final model (Model J) with all loss functions involved performs the best for all three measures for the validation set. It is slightly outperformed by the model with FMCL branch only (Model E) in AUC for the testing set. Model E comes second for the validation dataset, but not as generalizable for the testing set. Models A to C show the effectiveness if only one loss function is considered. Model D shows the result of no representation learning, and the performance is substantially worse comparing to Model J. The contribution of HF can be observed by comparing Models C with E, and Models I with J. Models F to H show the effect of missing a particular loss function.  

\begin{table*}[ht]
\renewcommand{\arraystretch}{1.25}
\caption{Ablation study on contrastive learning (stage 1 training)}
\centering
\begin{tabular}{c|cccc|ccc|ccc}
    \bottomrule
    \multirow{2}{*}{Model}&\multirow{2}{*}{\makecell[c]{$L_{intra}$}} & \multirow{2}{*}{\makecell[c]{{$L_{inter}$}}} & \multirow{2}{*}{$L_{FMCL}$} &\multirow{2}{*}{HF} &  \multicolumn{3}{c}{Validation set} & \multicolumn{3}{|c}{Testing set} \\ \cline{6-11}
        & & & & & AUC & F1 & Acc (\%)& AUC & F1 & Acc (\%)\\ \hline
        A & V & & &  & 0.8770 & 0.8237 & 82.73 & 0.8701 & 0.7869 & 79.39 \\ 
        B & & V & &  & 0.8701 & 0.8212 & 82.58 & 0.8613 & 0.7773 & 78.64 \\ 
        C & & & V &  & 0.8756 & 0.8153 &82.12 & 0.8682 & 0.7891 & 79.85  \\ 
        D & & & & V  & 0.8677 & 0.8163 & 81.82 & 0.8491 & 0.7662 & 77.12  \\  
        E & & & V & V & 0.8843 & 0.8271 & 83.03 & \textbf{0.8737} & 0.7895 & 79.70 \\ 
        F & V & V & & & 0.8612 & 0.8134 & 81.82 & 0.8680 & 0.7953 & 80.30 \\ 
        G & V & & V & V & 0.8772 & 0.8234 & 82.73 & 0.8689 & 0.7941 & 80.15 \\ 
        H & & V & V & V & 0.8685 & 0.8152 & 81.97 & 0.8656 & 0.7927 & 80.00 \\ 
        I & V & V & V & & 0.8635 & 0.8076 & 81.36 & 0.8547 & 0.7811 & 79.09 \\ 
        J & V & V & V & V & \textbf{0.8863} & \textbf{0.8304} & \textbf{83.48} & 0.8703 & \textbf{0.7968} & \textbf{80.45} \\
\toprule
\end{tabular}
\label{tableablation}
\end{table*}

We also investigate three different strategies of computing the final loss function for representation learning. The first one is to average $L_{intra}$, $L_{inter}$ and $L_{FMCL}$ so they all make equal contribution to the final loss. The second is to randomly select a loss for each epoch. The third approach involves randomly selecting a loss for each mini-batch. Table~\ref{table:traing strategy} shows the results. The first and the second strategies perform very similarly. The speculation for the more superior performance coming from the third strategy is the closer-to-equal probability of an individual loss function being optimized.   

\begin{table}[t]
\renewcommand{\arraystretch}{1.25}
\caption{Combined loss computation (stage 1 training)}
\centering
\begin{tabular}{c|ccc|ccc}
    \bottomrule
    \multirow{2}{*}{Training strategies}  & \multicolumn{3}{c}{Validation set} & \multicolumn{3}{|c}{Testing set} \\ \cline{2-7}
        &  AUC & F1 & Acc (\%) & AUC & F1 & Acc (\%)\\ \hline
        Averaging & 0.8760 & 0.8197 & 82.27 & 0.8623 & 0.7842 & 79.09 \\
        Randomly picked per epoch & 0.8760 & 0.8239 & 82.73 & 0.8642 & 0.7874 & 79.39 \\
        Randomly picked per mini-batch & \textbf{0.8863} & \textbf{0.8304} & \textbf{83.48} & \textbf{0.8703} & \textbf{0.7968} & \textbf{80.45} \\
\toprule
\end{tabular}
\label{table:traing strategy}
\end{table}

\subsection{Comparison with Single Modality Learning}
To study how much each modality can potentially contributes to the overall learning, we train baseline models, one for each modality: ResNet50 for ADC and DWI, and MLP for structured data. Table~\ref{table:single} shows the results.  We also train models without NIHSS variables (16 in total) to evaluate the performance of the system without the input of clinical variables with high inter-observer variability. NIHSS variables are also known to be the most effective variables driving the prediction of functional stroke outcome by MLP~\cite{c3}

\begin{table}[ht]
\renewcommand{\arraystretch}{1.25}
\caption{Single modality baseline results}
\centering
\begin{tabular}{lcccccc}
    \bottomrule
    & \multicolumn{3}{c}{Validation set} & \multicolumn{3}{c}{Testing set} \\ \cmidrule(lr){2-4} \cmidrule(lr){5-7}
        Model &  AUC & F1 & Acc (\%)& AUC & F1 & Acc (\%)\\ \midrule
        ResNet50 (ADC) & 0.7484 & 0.6960 & 70.15 & 0.7610 & 0.6932 & 70.30 \\
        ResNet50 (DWI) & 0.7832 & 0.7397 & 74.09 & 0.7552 & 0.6845 & 68.78 \\
        MLP (all clinical attributes) & 0.8726 & 0.8090 & 81.36 & 0.8569 & 0.7481 & 75.76 \\
        MLP (without NIHSS) & 0.8439 & 0.7828 & 78.94 & 0.8179 & 0.7447 & 75.30 \\ \hline
        Ours (all clinical attributes)& \textbf{0.8863} & \textbf{0.8304} & \textbf{83.48} & \textbf{0.8703} & \textbf{0.7968} & \textbf{80.45} \\
        Ours (without NIHSS)& 0.8401 & 0.7949 & 80.00 & 0.8408 & 0.7690 & 77.73 \\
        
\toprule
\end{tabular}
\label{table:single}
\end{table}

\section{Discussion}
Compared with other multimodal fusion learning networks, our proposed model performs the best in all three metrics. ~\cite{samak2020prediction, bacchi2020deep,hsu2021deep,huang2020multimodal} rate the second: to produce the final outcome, \cite{samak2020prediction, bacchi2020deep, hsu2021deep} perform simple concatenation for features from CNN and MLP before the FC layer, whereas \cite{huang2020multimodal} computes the average of the predicted probability from both CNN and MLP. \cite{hatami2022cnn}
performs the worst since the only structured attribute involved is only effective if the outcome from the CNN is sufficiently accurate.  

The fusion model makes significant improvement over CNN models for both ADC and DWI, which is in alignment with earlier studies~\cite{bacchi2020deep,hilbert2019data,zihni2020multimodal}. However, the improvement over MLP is considered incremental. \cite{zihni2020multimodal} came to a similar conclusion with whole-brain TOF-MRA modality: the CNN model achieves low performance (AUC:0.64), while MLP on clinical variables and multi-modal learning achieve comparable outcomes (AUCs:0.75 and 0.76, respectively). \cite{bacchi2020deep} claimed significant improvement from multi-modal learning trained on NCCT (AUC:0.75), due to much lower performance by CNN and MLP (AUCs:0.54 and 0.61, respectively). Interestingly, \cite{bacchi2020deep} achieves even higher performance in cross-modal learning with our dataset (AUCs:0.847). Our study confirms the superiority of diffusion-weighted MR imaging, compared with brain CT or other MR modalities, for functional outcome prediction. 

 The design of the fusion technique also plays an important role. \cite{hatami2022cnn} combines both perfusion and diffusion MRI modalities (5 in total), and clinical metadata, but achieves very similar performance as~\cite{zihni2020multimodal} in their own study of 119 patients (AUC: 0.77) and with our dataset as well (AUC:0.78). Table~\ref{table:single} also shows results of omitting NIHSS variables, which can lead to higher generalizability because NIHSS scores are subject to inter-observer variability. Our fusion model without NIHSS achieves comparable results as MLP using all 62 attributes. In other words, diffusion-weighted MRI can serve as a replacement for NIHSS assessment in predicting the 3-month functional outcome.

\begin{figure}[t]
\centering
\includegraphics[width=0.7\textwidth]{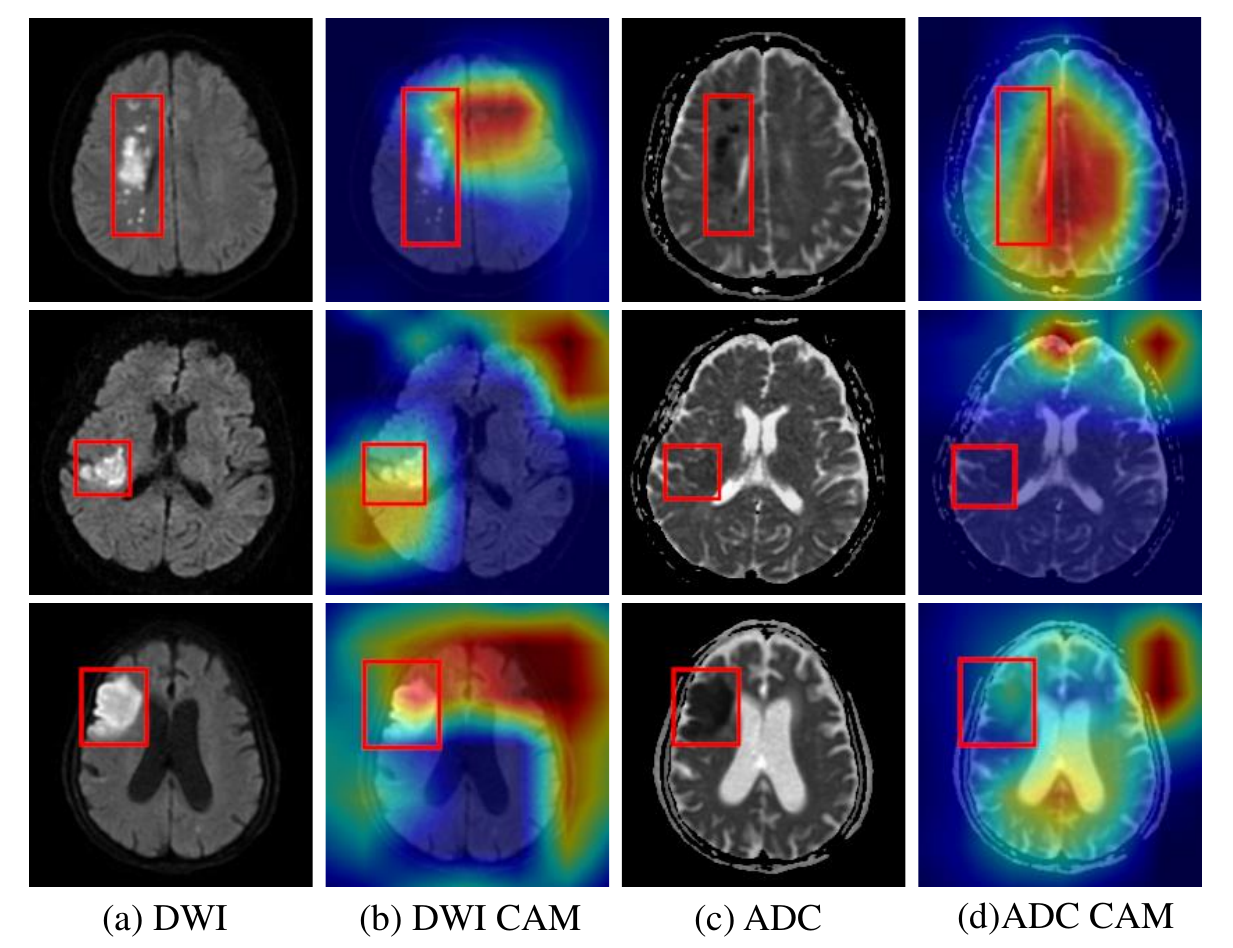}
\caption{Illustration of class activation maps for correctly predicted patients from the test set by CNN pre-trained with our representation learning. Our fusion model also succeeded on all 3 cases, but MLP failed. The red box highlights the affected region of the brain.}
\label{fig:cams} 
\end{figure}

We are interested to find out how ADC and DWI contribute to the prediction task in the joint framework, and if the decision is derived from clinically relevant evidence. The fusion model accurately predicted 6.3\% of the test set while MLP failed, but failed 1.6\% while MLP succeeded, for a net increase of 4.7\% in accuracy. For an image-based learning network, the explainability of a decision can be visualized by techniques, such as Gradient-weighted Class Activation Mapping (Grad-CAM~\cite{c4}), to identifies regions of interest in the input image that are important for making the prediction. Grad-CAM calculates the rate of change in the prediction of a target class regarding a change in the pixel/voxel location and display the analysis as a heatmap. Unfortunately, Grad-CAM cannot be applied to a fusion model because of the fusion layer. Instead we generate the Grad-CAM from CNN models pre-trained with our contrastive representation learning (stage 1). Fig.~\ref{fig:cams} shows 3 typical cases, one mild and two severe that only MLP on structured data failed. For all cases the CNN models focus on more than just the ischemic regions. Possible explanations include atrophy related to the patient’s age and abnormalities during image acquisition that are highly correlated with stroke severity. Other explainable techniques should be explored to properly visualize the interaction among the modals after the second stage of training.    

\section{Conclusion}
In this paper, we proposed a novel prognostic risk model for predicting the functional outcome 3 months after AIS tested with diffusion-weighted MRI imaging combined with clinical structured data. Our model applies representation learning in embeddings of individual modalities and in the fused multimodal embedding with hierarchical fusion to ensure equal weights for features of different granularities in the common feature embedding. Discriminative features are learned and applied to classification.

The proposed model outperforms existing risk-prediction fusion models for prognostication of AIS. Given a comprehensive health assessment of a patient, the prediction can be driven by the clinical variables, but the addition of diffusion-weighted MRI alone can further improve the accuracy of prediction, and can also replace NIHSS for better generalization. 

The study was initiated with a set of 4505 stroke patient records, but 1208 without 3-month mRS assessment and discarded for the current study for lacking ground truth. As the next step, we will expand our model to also leverage data without label for semi-supervised learning to further improve generalization of the model. Different post-hoc explainability methods for a fusion model should also be explored to establish the connection between DL features and imaging properties already identified with clinical relevance to confirm the reliability of the fusion model as a clinical decision support system.

\section*{Acknowledgment}
This work was partially supported by grants PSC-CUNY Research Award 65406-00 53, CYCH-CCU-2022-14, and NSTC 112-2221-E-194-034.

\bibliographystyle{unsrtnat}







\end{document}